\pdfoutput=1
\documentclass[letterpaper]{article}
\usepackage[preprint]{aaai2027}
\usepackage[hyphens]{url}
\usepackage{graphicx}
\usepackage{natbib}
\usepackage{caption}
\usepackage{booktabs}
\usepackage{amsmath}
\usepackage{amssymb}
\usepackage{xcolor}
\usepackage{colortbl}

\newcommand{\model}{MR-MoL}

\title{Multi-Granular Rationale-Guided Molecular LLM for Property Prediction}

\author{
    Junwoo Park\textsuperscript{\rm 1},
    Minyoung Shin\textsuperscript{\rm 1},
    Cheol Soon Lee\textsuperscript{\rm 2},
    Sujee Lee\textsuperscript{\rm 1}\corresponding
}
\affiliations{
    \textsuperscript{\rm 1}Sungkyunkwan University,
    \textsuperscript{\rm 2}Korea University\\
    jw0528@g.skku.edu, ericashin1@g.skku.edu,
    xtal21@korea.ac.kr, sujeelee@skku.edu
}

\begin{document}
\maketitle

\begin{abstract}
Large language models (LLMs) are widely applied across chemical tasks, such as molecular property prediction, which underpins drug discovery. Molecular LLMs represent a molecule through several modalities, notably a 1D SMILES sequence or a 2D molecular graph. Both encode molecular information implicitly, so the contribution of individual substructures remains opaque. Retrieval and augmentation methods add context, but from external sources. However, the cues chemists reason over are the internal substructures that drive a property up or down. We propose MR-MoL, a multi-granular rationale-guided molecular LLM that supplies this evidence directly. A fine-tuned GNN scores each substructure through masking, and the most influential ones are serialized as a ranked, direction-tagged rationale that the LLM reads alongside the SMILES sequence and molecular graph. The rationale spans three levels of granularity: Murcko scaffolds with their side chains, BRICS fragments, and functional groups. This is, to our knowledge, the first method to expose GNN-derived attributions to an LLM as evidence for property prediction. On eight MoleculeNet tasks, MR-MoL achieves the best overall results among generalist models and narrows the gap to specialist models tuned for each task. Five diagnostics further confirm that the model reads the rationale rather than merely benefiting from its presence. Its direction, rank, and substructure each shape the prediction, and its attributions reproduce known structure-property relationships.
\end{abstract}

\begin{links}
\link{Code}{https://github.com/skku-aihclab/MR-MoL}
\end{links}

\section{Introduction}

Molecular property prediction underpins drug discovery, toxicity screening, and solubility design. Deep learning models for this task have traditionally operated directly on diverse molecular representations: 2D molecular graphs~\cite{xia2023mole, wang2022molecular, liu2022pretraining}, 1D Simplified Molecular-Input Line-Entry System (SMILES) sequences~\cite{ross2022large, chithrananda2020chemberta}, and 3D conformers~\cite{zhou2023unimol, schutt2018schnet}. Such models remain among the strongest predictors on standardized benchmarks such as MoleculeNet~\cite{wu2018moleculenet}, but are typically trained or fine-tuned per task. 

More recently, large language models (LLMs) have been adapted to molecular property prediction. Applying an LLM to a molecule first requires casting its representation into a form the model can consume, and two routes have emerged. The first route keeps everything textual. The molecule is serialized as a string and fed to the LLM as text tokens~\cite{yu2024llasmol, fang2024molinstructions}. The second route attaches a graph encoder and aligns it to the LLM through a projector, so that the molecule enters the prompt as continuous embeddings. Some of these models are fine-tuned per task~\cite{liu2023molca, chen2024hight}, while others share a single model over many chemical tasks~\cite{park2024llamo, kim2026mol}.

Yet these molecular LLMs leave a common gap. Since modalities such as 2D graph and 1D SMILES representations encode molecular information implicitly, the contribution of the molecule's substructures remains opaque~\cite{wellawatte2023perspective, jimenez2020drug, wu2023chemistry}. Adding retrieval or knowledge graphs to the prompt can supply extra context at the molecule level~\cite{xian2025molrag, lee2026rag}, but none of these channels exposes the importance of the current molecule's substructures in a form the model can read. In practice, this is the evidence chemists rely on, namely a functional group or fragment that elevates or suppresses a property. Motivated by this gap, we equip the LLM with explicit, per-substructure evidence specific to the input molecule.

We propose \model{}, a \textbf{M}ulti-granular \textbf{R}ationale-Guided \textbf{Mo}lecular \textbf{L}LM. The idea is simple. We let a fine-tuned graph neural network (GNN) score each substructure's contribution to its prediction~\cite{wu2023chemistry}, and give this to the LLM as a rationale, a ranked and direction-tagged list indicating whether each substructure pushes the GNN's prediction higher or lower. We serialize it as text, so the LLM reads it directly alongside SMILES sequences and molecular graph representations. Specifically, we draw the rationale from three views at different structural granularities: Murcko scaffolds together with their side chains~\cite{bemis1996properties}, retrosynthetically motivated BRICS fragments~\cite{degen2008art}, and chemically named functional groups.

This design supports two claims. First, the rationale should improve predictive performance, since structural cues are known to matter in molecular property prediction~\cite{fang2023knowledge, zhang2021motif, wu2023chemistry}. Second, the LLM should use rationales as directional and ranked evidence, not as inert filler. We evaluate the first claim against seven specialist models, comprising five GNN-based models and two molecular LLMs that are fine-tuned per task, and five generalist molecular LLMs that share a single model across tasks. We test the second claim with five diagnostics, four behavioral interventions that edit the rationale and one chemical check on whether its attributions reproduce known structure-property relationships. 

Our contributions are as follows.
\begin{itemize}
    \item We introduce \model{}, a multi-task molecular LLM for property prediction. To our knowledge, it is the first to feed GNN-derived attributions into the prompt as evidence.
    \item Building on substructure masking, we formulate a rationale that decomposes per-substructure attribution into three complementary views and exposes it as direction-tagged, ranked textual evidence.
    \item We benchmark \model{} on eight MoleculeNet tasks with ablations, and design diagnostics that show the rationale acts as effective evidence.
\end{itemize}

\section{Related Work}

\paragraph{Instruction-Tuned Molecular Language Models.}
A growing body of molecular LLMs is instruction-tuned on molecule-text pairs to follow natural-language task descriptions~\cite{wei2022finetuned, ouyang2022training}. LlaSMol~\cite{yu2024llasmol} and nach0~\cite{livne2024nach0} tune LLMs on large chemistry instruction corpora, representing molecules as SMILES sequences. The molecule reaches the LLM through this textual prompt alone. The other group attaches a modality-specific encoder and aligns it to the LLM through a projector. MolCA~\cite{liu2023molca}, LLaMo~\cite{park2024llamo}, and InstructMol~\cite{cao2025instructmol} project a molecular graph into the LLM, while others enrich this representation with a 2D image~\cite{liu2024git}, hierarchical motif-level graph tokens~\cite{chen2024hight}, or a 3D conformer~\cite{li2024towards, kim2026mol}. Whether molecular information enters as text or as projected tokens, neither path makes the substructure-level signal legible to the model.

\paragraph{Molecular GNNs and Explainability.}
GNNs treat a molecule as a graph of atoms and bonds, learning representations by message passing between neighboring atoms~\cite{scarselli2008graph, gilmer2017neural}. To learn representations that transfer across tasks, self-supervised pretraining has become the dominant approach~\cite{Hu*2020Strategies, wang2022molecular, liu2022pretraining}. Such pretrained GNNs are strong predictors, yet the effect of an individual substructure stays hidden within their representations. To expose it, a parallel line of work attributes predictions back to subgraphs, with GNNExplainer~\cite{ying2019gnnexplainer} ranking individual nodes and edges and SubgraphX~\cite{yuan2021explainability} scoring whole connected subgraphs. Following \citet{wu2023chemistry}, we adopt substructure masking, which scores each fragment by the prediction change caused when it is removed. Such interpretability is widely viewed as essential for bridging black-box predictors and human-understandable chemistry~\cite{jimenez2020drug, wellawatte2023perspective}.

\paragraph{Knowledge-Augmented Molecular Prediction.}
A separate line of work augments property prediction with auxiliary knowledge. For instance, LLM-MPP~\cite{jin2025effective} and MolProphecy~\cite{zhao2025molprophecy} encode LLM-generated descriptions or chemical knowledge and fuse them with molecular features. Other approaches focus on retrieving supporting evidence; MolRAG~\cite{xian2025molrag} draws on analogous molecules, while CLADD~\cite{lee2026rag} employs collaborative LLM agents that query knowledge bases. Separately, KANO~\cite{fang2023knowledge} builds chemical priors into the GNN by coupling graph contrastive learning with a functional group knowledge graph. Across these methods, the supplied evidence is external or holistic, rather than identifying how specific substructures influence the target property.

\section{Method}

\model{} takes four inputs: a task instruction $I$, a 1D SMILES sequence $S$, a 2D molecular graph $G$, and a multi-granular rationale $R$. The output $Y$ is the answer to the task, produced as text.

\begin{figure*}[t]
    \centering
    \includegraphics[width=0.95\textwidth]{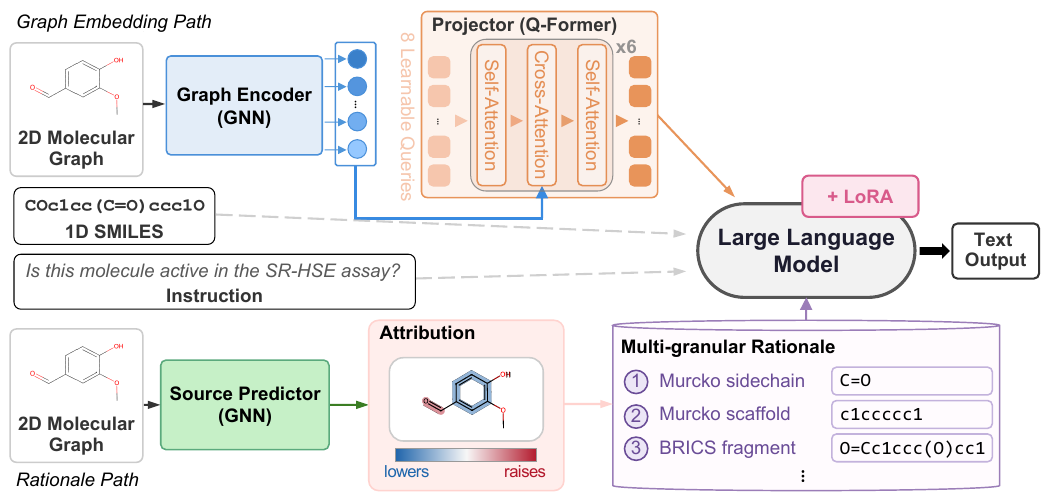}
    \caption{Overview of \model{}. The graph embedding path encodes a 2D molecular graph through a graph encoder and the projector, yielding molecular tokens. The rationale path applies substructure masking to a source predictor, scoring substructures by their attribution. The top-scoring substructures form the items of a multi-granular rationale that is provided to the LLM together with the instruction, SMILES, and molecular tokens, and the model generates the answer to the task.}
    \label{fig:overall}
\end{figure*}

Two complementary information paths feed the LLM. The \textbf{graph embedding path} turns $G$ into molecular tokens through a GNN encoder and a projector. The \textbf{rationale path} masks substructures against a fine-tuned GNN and serializes the most influential ones as ranked items. The LLM conditions on the instruction, SMILES sequence, molecular tokens, and rationale to produce $Y$, as shown in Figure~\ref{fig:overall}.

We train \model{} in two stages. Stage~1 performs molecular graph-language alignment for the graph embedding path. Stage~2 performs multi-task, rationale-guided instruction tuning for property prediction.

\subsection{Molecular Graph-Language Alignment}

The projector module for graph-language alignment follows the Q-Former design~\cite{li2023blip}, which has been adopted for grounding molecular graphs in LLMs. We initialize the GNN encoder and Q-Former with weights pre-trained on molecule-text data~\cite{liu2023molca}, and then train them with a newly added linear projection during Stage~1 to align the molecular tokens with the embedding space of our LLM.

Given $G=(V,E)$, the encoder $E_{\phi}$ with parameters $\phi$ produces atom-level features:
\begin{equation}
    H_G = E_{\phi}(G), \qquad H_G \in \mathbb{R}^{N \times d_g},
\end{equation}
where $N$ is the number of atoms and $d_g$ is the graph hidden dimension. The Q-Former with parameters $\psi$ uses $K$ learnable queries $Q \in \mathbb{R}^{K \times d_q}$ and cross-attends to $H_G$, where $d_q$ is the Q-Former hidden dimension:
\begin{equation}
    Z = \mathrm{QFormer}_{\psi}(Q, H_G), \qquad Z \in \mathbb{R}^{K \times d_q}.
\end{equation}
A linear projection $W_p \in \mathbb{R}^{d_q \times d_{\ell}}$ then maps the query embeddings $Z$ to the LLM hidden dimension $d_{\ell}$:
\begin{equation}
    M = Z W_p, \qquad M \in \mathbb{R}^{K \times d_{\ell}}.
\end{equation}
These molecular tokens $M$ are inserted into the LLM input sequence.

\subsection{Ranked Multi-Granular Rationale}

The rationale is task-conditioned textual evidence built from a fine-tuned GNN. We decompose the molecule into candidate substructures, score each by its attribution~\cite{wu2023chemistry}, and serialize the top ones as ranked text.

\paragraph{Multi-granular decomposition.} For each molecule, we decompose the structure at three granularities, each realized as a view. The \textbf{Murcko view} splits the molecule into its scaffold and side chains through Murcko decomposition~\cite{bemis1996properties}. The \textbf{BRICS view} provides retrosynthetically motivated fragments with attachment information~\cite{degen2008art}. The \textbf{functional group view} lists chemically named local motifs such as amide and carbonyl. The three views are complementary, spanning a range of structural granularity from molecular scaffolds down to local motifs.

\paragraph{Attribution and ranking.} Let $f_t$ denote the fine-tuned source predictor for task $t$, whose output $f_t(G)$ is the positive-class score for classification or the predicted value for regression. Let $\mathcal{U}(G)$ be the candidate substructures pooled from the three views, indexed by $j$. For each substructure $u_j \in \mathcal{U}(G)$, the attribution is
\begin{equation}
    a_{t,j} = f_t(G) - f_t(G \setminus u_j),
    \label{eq:attribution}
\end{equation}
where $G \setminus u_j$ is the graph with $u_j$ masked. A positive $a_{t,j}$ means that removing $u_j$ lowers the prediction, so the substructure drives the prediction up; a negative $a_{t,j}$ indicates the reverse. We rank substructures within each molecule by $|a_{t,j}|$ and keep the top five, those producing the largest change in the source prediction.

\paragraph{Serialization.} The rationale $R$ lists the top five substructures as ranked items. Each item carries three fields, the view type, the substructure, and the effect. The substructure is written as scaffold or side chain SMILES for the Murcko view, as fragment SMILES with attachment points for the BRICS view, and as the group name for the functional group view. The effect field tags the item as \textit{toward higher} or \textit{toward lower}, applied to the predicted probability of the target property for classification and the predicted value for regression. This wording is an attribution about the source predictor, not a chemical mechanism claim. A complete serialized input is given in Appendix~B.

\subsection{Training}

\paragraph{Two-stage training.} Both stages cast learning as instruction following over samples in a chat format, where the molecular tokens replace the molecule-embedding placeholders and the model is trained to generate the target $Y$. Serving different purposes, the two stages differ in the target $Y$ and the presence of the rationale $R$. In Stage~1, each sample takes the form $(I,S,G,Y)$ and the target $Y$ is a molecule description. In Stage~2, each sample takes the form $(I,S,G,R,Y)$, adding the rationale $R$, and $Y$ is the answer to the property prediction task. Stage~1 makes the molecular tokens legible to the LLM, a grounding step where the rationale plays no role. Stage~2 then inherits these grounded tokens and adds the rationale as task-conditioned evidence for property prediction.

\paragraph{Objective.} We optimize next-token prediction, where the prompt tokens are excluded from the loss and only the answer tokens contribute. Writing $Y=(y_1,\dots,y_T)$ for the $T$ answer tokens and $\widetilde{X}$ for the embedded input sequence, the answer likelihood factorizes as
\begin{equation}
    p_{\theta}(Y \mid G,S,I,R)
    =
    \prod_{m=1}^{T}
    p_{\theta}(y_m \mid y_{<m}, \widetilde{X}),
\end{equation}
and the training loss is its negative log-likelihood,
\begin{equation}
    \mathcal{L}_{\mathrm{LM}}
    =
    -\sum_{m=1}^{T}
    \log p_{\theta}(y_m \mid y_{<m}, \widetilde{X}).
\end{equation}
Stage~1 minimizes the loss over the molecule-text data, whose targets are descriptions, and Stage~2 over the multi-task instruction data, whose targets are property prediction answers. Each stage selects its checkpoint by validation loss.

\section{Experiments}

\subsection{Datasets and Preprocessing}

Our data is organized by training stage. Stage~1 draws molecule-text pairs from PubChem324k~\cite{liu2023molca}, DrugBank~\cite{knox2024drugbank}, Mol-Instructions~\cite{fang2024molinstructions}, and ChEBI-20~\cite{edwards2021text2mol, hastings2012chebi}. To ensure quality, we keep only pairs whose description spans at least 30 non-whitespace characters, and we drop those duplicated across different molecules. Molecules shared across sources are also deduplicated. This filtering yields 89,919 samples. The instructions for this stage follow Mol-Instructions. We hold out the ChEBI-20 test split as validation data for Stage~1 checkpoint selection.

Stage~2 covers eight MoleculeNet datasets~\cite{wu2018moleculenet}, six classification and two regression, where ClinTox, SIDER, and Tox21 are multi-label. These span physiological and biophysical classification, such as toxicity (ClinTox, Tox21, SIDER), blood-brain barrier permeability (BBBP), and bioactivity (BACE, HIV), together with physicochemical regression on aqueous solubility (ESOL) and lipophilicity (Lipo). Each dataset is divided 8:1:1 into train, validation, and test sets using a scaffold split, and the per-task sizes are reported in Appendix~A. We build the Stage~2 instructions on SMolInstruct~\cite{yu2024llasmol}, supplementing the uncovered datasets with Claude Opus 4.6~\cite{anthropic2026opus46}, and evaluate the molecular LLMs with the same instructions.

All SMILES are canonicalized with \citet{rdkit}, and molecules shared between stages are removed.

\subsection{Baselines}
\label{sec:comparison-protocol}

We compare \model{} against two families of baselines, specialist models and generalist models for property prediction.

\paragraph{Specialist.} We define specialist models as those fine-tuned per task. We include seven such baselines, spanning GNN-based models and molecular LLMs. Here a GNN-based model encodes the molecular graph with a GNN and outputs the property directly. These models differ in their pretraining strategy. MolCLR~\cite{wang2022molecular} and GraphMVP~\cite{liu2022pretraining} use contrastive learning, and Mole-BERT~\cite{xia2023mole} combines it with masked atom modeling. MGSSL~\cite{zhang2021motif} is motif-generative, and KANO~\cite{fang2023knowledge} augments pretraining with a chemical knowledge graph. Among the molecular LLMs, 3D-MolT5~\cite{pei20253d} combines 1D and 3D representations, while HIGHT~\cite{chen2024hight} uses hierarchical graph tokenization.

\paragraph{Generalist.} We define generalist models as molecular LLMs that are instruction-tuned and applied across many tasks. We include five generalist baselines, grouped by how they represent molecules. LlaSMol~\cite{yu2024llasmol}, ChemDFM~\cite{zhao2024chemdfm}, nach0~\cite{livne2024nach0}, and MolecularGPT~\cite{liu2024moleculargpt} are string-based models that process molecules as 1D sequences such as SMILES or SELFIES. In contrast, GIMLET~\cite{zhao2023gimlet} is a graph-text model that directly encodes 2D molecular graphs.

\subsection{Implementation Details}

We use Llama-3.1-8B-Instruct~\cite{grattafiori2024llama} as the LLM. For the graph embedding path, the GNN encoder and Q-Former are initialized from MolCA's~\cite{liu2023molca} stage-1 pre-trained weights. The Q-Former uses eight learnable query tokens and is followed by a projection layer. In Stage~1, we freeze the LLM and train the GNN encoder and the projector. In Stage~2, the GNN encoder stays frozen, and we adapt the LLM with Low-Rank Adaptation (LoRA)~\cite{hu2022lora} by training the adapters together with the projector. For each task, we fine-tune a separate GNN initialized from Mole-BERT~\cite{xia2023mole} to generate rationales. We optimize with AdamW~\cite{loshchilov2018decoupled} under a cosine schedule with linear warmup, and provide the full training and optimization details in Appendix~A.

Every component of our method is run three times under the same split for statistical robustness. This covers both the GNNs that generate the rationales and all \model{} variants in Table~\ref{tab:ablation-core}. Diagnostic experiments perturb only the inference-time input of \model{} on the test split. Every other baseline is reproduced in a single run. We fine-tune specialist models on our training split, whereas generalist models are evaluated from their publicly available checkpoints.

\subsection{Evaluation Metrics}
We report ROC-AUC for classification and RMSE for regression as the main metrics, following MoleculeNet's evaluation protocol~\cite{wu2018moleculenet}. On multi-label datasets, ROC-AUC is computed per endpoint and averaged across endpoints.

Unlike GNN-based models, which output the score directly, molecular LLMs including \model{} require reading the score from the generated output. In classification, we apply a two-way softmax over the logits of the positive and negative answer tokens (e.g., ``Yes'' and ``No''). With logits $z_{\mathrm{Yes}}$ and $z_{\mathrm{No}}$, the positive-class score is
\begin{equation}
    P(\mathrm{Yes})
    =
    \frac{\exp(z_{\mathrm{Yes}})}{\exp(z_{\mathrm{Yes}})+\exp(z_{\mathrm{No}})},
\end{equation}
which we use for ROC-AUC. In regression, we parse the first valid number, and exclude outputs without a parseable number from the computation.

For the direction diagnostic of \model{}, we additionally report the Matthews correlation coefficient (MCC) for classification and the mean error (ME) for regression.

\begin{table*}[t]
\centering
\small
\begin{tabular}{llcccccccc}
\toprule
 & & \multicolumn{6}{c}{Classification (ROC-AUC\,$\uparrow$)} & \multicolumn{2}{c}{Regression (RMSE\,$\downarrow$)} \\
\cmidrule(lr){3-8} \cmidrule(lr){9-10}
Model & Base LLM & BACE & BBBP & ClinTox & HIV & SIDER & Tox21 & ESOL & Lipo \\
\midrule
\rowcolor{gray!20}\multicolumn{10}{l}{\textit{\textbf{Specialist} -- GNN-based models}} \\
MolCLR    & -- & 81.0 & 73.1 & 86.3 & 75.1 & 61.4 & 73.5 & 1.237 & \underline{\textbf{0.711}} \\
MGSSL     & -- & \underline{\textbf{86.6}} & 72.8 & 71.0 & 72.7 & 60.0 & 72.8 & 1.303 & 0.784 \\
GraphMVP  & -- & 78.2 & 67.5 & 67.6 & 75.4 & 61.9 & 73.2 & 1.105 & 0.729 \\
KANO      & -- & 80.7 & 70.1 & \underline{\textbf{89.1}} & 72.5 & 59.9 & 73.6 & \underline{\textbf{0.834}} & 0.712 \\
Mole-BERT & -- & 80.9 & \underline{\textbf{73.2}} & 75.4 & 75.4 & \textbf{62.2} & \underline{\textbf{77.3}} & 1.102 & 0.731 \\
\midrule
\rowcolor{gray!20}\multicolumn{10}{l}{\textit{\textbf{Specialist} -- Molecular LLMs}} \\
3D-MolT5 & T5-v1.1-base   & 82.0 & 70.1 & 85.9 & 64.6 & 55.3 & 67.5 & 1.260 & 0.894 \\
HIGHT    & Vicuna-7B-v1.3 & 78.9 & 68.9 & 56.1 & \textbf{76.5} & 58.9 & 72.9 & 1.854 & 0.983 \\
\midrule
\rowcolor{gray!20}\multicolumn{10}{l}{\textit{\textbf{Generalist} -- Molecular LLMs}} \\
GIMLET       & T5-small         & 59.0 & 52.6 & 49.9 & 49.5 & 46.8 & 56.3 & 7.557 & 1.545 \\
nach0        & T5-base          & 63.6 & 71.0 & 49.0 & \textbf{76.1} & 52.8 & 50.7 & 3.745 & \textbf{0.874} \\
ChemDFM      & LLaMA-13B        & 70.8 & 50.0 & 34.0 & 69.1 & 53.0 & 57.3 & 17.149 & 3.497 \\
LlaSMol      & Mistral-7B       & 56.2 & 64.9 & 53.0 & 68.2 & 56.4 & 51.2 & 4.306 & 0.879 \\
MolecularGPT & Llama-2-7B-chat  & 61.6 & 55.7 & 52.9 & 69.8 & 53.6 & 57.0 & 12.812 & 1.149 \\
\model{} (ours) & Llama-3.1-8B-Instruct & \textbf{82.6} & \textbf{71.8} & \textbf{73.8} & 72.0 & \underline{\textbf{63.3}} & \textbf{74.2} & \textbf{1.210} & 0.919 \\
\bottomrule
\end{tabular}
\caption{Results on eight MoleculeNet property prediction tasks. Classification is reported as ROC-AUC and regression as RMSE. Within the Specialist and Generalist groups, the best score per task is in \textbf{bold} and the overall best across groups is \underline{underlined}. GNN-based models have no base LLM and are marked --.}
\label{tab:main}
\end{table*}

\begin{table*}[t]
\centering
\small
\setlength{\tabcolsep}{5pt}
\begin{tabular}{lcccccccc}
\toprule
Variant & BACE\,($\uparrow$) & BBBP\,($\uparrow$) & ClinTox\,($\uparrow$) & HIV\,($\uparrow$) & SIDER\,($\uparrow$) & Tox21\,($\uparrow$) & ESOL\,($\downarrow$) & Lipo\,($\downarrow$) \\
\midrule
w/o $G$, $R$ & $74.9_{\pm 2.9}$ & $70.1_{\pm 1.5}$ & $63.9_{\pm 3.5}$ & $74.3_{\pm 1.0}$ & $61.0_{\pm 1.0}$ & $71.8_{\pm 1.0}$ & $1.892_{\pm 0.037}$ & $1.084_{\pm 0.025}$ \\
w/o $G$ & $78.3_{\pm 1.5}$ & $71.0_{\pm 1.5}$ & $66.3_{\pm 3.3}$ & $\mathbf{74.4}_{\pm 0.6}$ & $60.6_{\pm 1.4}$ & $73.2_{\pm 1.1}$ & $1.547_{\pm 0.151}$ & $1.089_{\pm 0.015}$ \\
w/o $R$     & $77.9_{\pm 2.3}$ & $68.0_{\pm 4.1}$ & $64.2_{\pm 5.9}$ & $73.2_{\pm 1.8}$ & $61.9_{\pm 0.4}$ & $72.1_{\pm 0.9}$ & $1.425_{\pm 0.010}$ & $1.027_{\pm 0.025}$ \\
\model{} & $\mathbf{82.6}_{\pm 0.5}$ & $\mathbf{71.8}_{\pm 2.8}$ & $\mathbf{73.8}_{\pm 1.4}$ & $72.0_{\pm 1.4}$ & $\mathbf{63.3}_{\pm 1.3}$ & $\mathbf{74.2}_{\pm 0.6}$ & $\mathbf{1.210}_{\pm 0.106}$ & $\mathbf{0.919}_{\pm 0.006}$ \\
\bottomrule
\end{tabular}
\caption{Ablation of \model{} on molecular graph $G$ and rationale $R$. Classification is reported as ROC-AUC and regression as RMSE. All variants report mean $\pm$ std. Best per task in bold.}
\label{tab:ablation-core}
\end{table*}

\section{Results and Analysis}

We organize our analysis around the two claims of the introduction, improved prediction and effective evidence use.

\subsection{Property Prediction Results}
Table~\ref{tab:main} reports the property prediction results, and Table~\ref{tab:ablation-core} reports the ablation. Baselines in Table~\ref{tab:main} are grouped into specialist and generalist families. For the ablation, we vary the two channels \model{} adds to the LLM, the molecular graph $G$ and the rationale $R$. \textbf{W/o $G,R$} removes both channels, leaving only the instruction and SMILES, and \textbf{w/o $G$} removes only the graph. Neither uses the Stage~1 checkpoint, as the graph path is absent. \textbf{W/o $R$} removes the rationale.

\paragraph{Comparison with baselines.} Among the generalist models, \model{} outperforms every baseline on six of the eight tasks, and the margin is often substantial. It leads by more than $11$ ROC-AUC points on BACE and by nearly $7$ on SIDER. The advantage carries over to regression, where every other generalist exceeds $3.7$ RMSE on ESOL against our $1.210$. We attribute this to the structural evidence \model{} receives alongside the 1D and 2D molecule representations. The two exceptions are HIV and Lipo, where nach0 leads.

More notably, \model{} closes much of the gap to the specialist models, which are fine-tuned per task. It surpasses both molecular LLMs, 3D-MolT5 and HIGHT, on four of the six classification tasks and on ESOL, and on SIDER it achieves the best result across all baselines. Elsewhere in classification it trails the strongest specialist by at most $4.5$ points, with the sole exception of ClinTox, where KANO keeps a clear lead. In regression the strongest GNN-based models retain an advantage on both tasks, as generating numbers remains harder for an LLM. We report output validity, the fraction of parseable answers for molecular LLMs, in Appendix~C.

\paragraph{Ablation.} As shown in Table~\ref{tab:ablation-core}, removing the rationale lowers performance on seven of the eight tasks, with the largest classification drops on ClinTox and BACE, confirming that the rationale carries information the model uses. Removing the graph also hurts most tasks, though typically by a smaller margin on classification. The two channels contribute unequally across tasks, with the rationale mattering more on classification and the graph being the larger contributor on the regression tasks. This pattern indicates that the rationale complements the molecular graph, and that the model draws on whichever channel is more informative for a given task instead of relying on one alone. The sole exception is HIV, where neither added channel helps and the w/o $G$,$R$ variant already suffices. Notably, w/o $G$,$R$ already surpasses the generalist baselines in Table~\ref{tab:main} on most tasks, since it shares our strong LLM backbone and multi-task instruction tuning. The gains of \model{} over this variant thus isolate the contribution of the $G$ and $R$ channels. Extended metrics, including F1, MCC, and accuracy, appear in Appendix~C.

\subsection{Rationale Diagnostics}
\label{sec:diagnostic}
 
A rationale that raises performance matters only if the model reads it rather than benefiting from its mere presence. We probe this from two sides. Behaviorally, we intervene on the rationale, by flipping a direction, removing an item, or replacing a substructure, and measure how the prediction moves on four datasets (BACE, BBBP, ESOL, and Lipo), then trace how the rationale shifts the answer on individual molecules. Chemically, we ask whether the attributions the rationale carries reproduce known structure-property relationships. All diagnostics use a single checkpoint of \model{}, so the unperturbed scores differ slightly from Table~\ref{tab:main}.
 
\paragraph{Direction sensitivity.} We flip only the direction tag in each item, so that \emph{toward higher} becomes \emph{toward lower} and the reverse, while leaving the substructure and its rank intact. A model that reads the tags should move its prediction. Table~\ref{tab:signflip} reports the effect. On both classification datasets the ROC-AUC falls below chance and the MCC flips sign, so \model{} does not merely lose the signal but follows the inverted tag. The reversal is not total, as the MCC magnitude shrinks rather than mirroring the original, indicating that \model{} weighs the direction tag together with other signals. On both regression datasets the RMSE rises while the mean error changes sign, so the direction also steers the numerical prediction.
 
\begin{table}[t]
\centering
\small
\setlength{\tabcolsep}{4pt}
\begin{tabular}{llcc}
\toprule
Dataset & Metric & Original & Flipped \\
\midrule
BACE & ROC-AUC (MCC) & $83.2$ ($+0.49$) & $33.4$ ($-0.30$) \\
BBBP & ROC-AUC (MCC) & $71.3$ ($+0.41$) & $29.2$ ($-0.29$) \\
\midrule
ESOL & RMSE (ME) & $1.332$ ($-0.43$) & $1.371$ ($+0.19$) \\
Lipo & RMSE (ME) & $0.928$ ($+0.10$) & $1.031$ ($-0.21$) \\
\bottomrule
\end{tabular}
\caption{Direction flipping. Each rationale item's direction tag is inverted. ``Original'' and ``Flipped'' report performance before and after the flip.}
\label{tab:signflip}
\end{table}
 
\paragraph{Rank sensitivity.} Rationale items are ranked by attribution magnitude. For each molecule $i$ we remove the rank-1 item and measure the prediction shift
\begin{equation}
\Delta_{\mathrm{top},i} =
\begin{cases}
P_{\mathrm{full}}(\mathrm{Yes})_i - P_{-\mathrm{top}}(\mathrm{Yes})_i & \text{classification},\\
\hat{y}_{\mathrm{full},i} - \hat{y}_{-\mathrm{top},i} & \text{regression},
\end{cases}
\end{equation}
where $P(\mathrm{Yes})$ is the positive-class probability and $\hat{y}$ is the predicted numeric value. The subscript $\mathrm{full}$ denotes the prediction given the entire rationale, and $-\mathrm{top}$ the prediction after removing the rank-1 item. We define $\Delta_{\mathrm{rand},i}$ in the same way, removing instead one item drawn at random from the remaining items below rank 1. We report $|\Delta_{\mathrm{top}}|$ and $|\Delta_{\mathrm{rand}}|$, the mean of $|\Delta_{\mathrm{top},i}|$ and $|\Delta_{\mathrm{rand},i}|$ over all molecules. The rank-1 item is the one whose masking most changed the source prediction, so a model that treats rank as importance should react most when it is removed. As Table~\ref{tab:rank} shows, $|\Delta_{\mathrm{top}}|$ exceeds $|\Delta_{\mathrm{rand}}|$ by $1.6$ to $6.2$ times across the four datasets. A paired $t$-test confirms the gap is significant on every dataset, with all $p<10^{-5}$.
 
\begin{table}[t]
\centering
\small
\begin{tabular}{lcccc}
\toprule
Dataset & $|\Delta_{\mathrm{top}}|$ & $|\Delta_{\mathrm{rand}}|$ & ratio & $p$ \\
\midrule
BACE & 0.276 & 0.049 & $5.6\times$ & $2\times10^{-27}$ \\
BBBP & 0.263 & 0.042 & $6.2\times$ & $4\times10^{-22}$ \\
\midrule
ESOL & 0.539 & 0.192 & $2.8\times$ & $7\times10^{-7}$ \\
Lipo & 0.321 & 0.196 & $1.6\times$ & $5\times10^{-6}$ \\
\bottomrule
\end{tabular}
\caption{Rank-1 versus random lower-rank removal. ``ratio'' is $|\Delta_{\mathrm{top}}|/|\Delta_{\mathrm{rand}}|$; $p$ is from a paired $t$-test between the two.}
\label{tab:rank}
\end{table}
 
\paragraph{Substructure sensitivity.} The first two tests vary the direction and the rank, but leave each item's substructure unchanged. Here we replace only its substructure with a random one of the same view drawn from the test pool. Table~\ref{tab:shuffle} confirms a drop on every dataset. This shuffling lowers ROC-AUC by $2$ to $4$ points on BACE and BBBP and raises RMSE on ESOL and Lipo. The model therefore depends on the specific structure each item names. Together, these three interventions show that \model{} reads every aspect of a rationale, its direction, its rank, and its substructures.
 
\begin{table}[tb]
\centering
\small
\begin{tabular}{lcccc}
\toprule
Rationale & BACE ($\uparrow$) & BBBP ($\uparrow$) & ESOL ($\downarrow$) & Lipo ($\downarrow$) \\
\midrule
Shuffled & $80.4$ & $67.5$ & $1.386$ & $1.020$ \\
Original & $83.2$ & $71.3$ & $1.332$ & $0.928$ \\
\bottomrule
\end{tabular}
\caption{Substructure shuffling. ``Original'' is the intact rationale, while ``Shuffled'' replaces each item's substructure with a random one. Classification is reported as ROC-AUC and regression as RMSE.}
\label{tab:shuffle}
\end{table}

\begin{table}[tb]
\centering
\small
\begin{tabular}{lcc}
\toprule
Functional group & ESOL & Lipo \\
\midrule
Primary sulfonamide & $+1.06$ (100\%) & $-0.92$ (100\%) \\
Methyl amide & $+1.05$ (100\%) & $-0.67$ (99\%) \\
Ester & $+0.62$ (91\%) & $-0.24$ (88\%) \\
Ketone & $+0.61$ (96\%) & $-0.37$ (99\%) \\
Amide & $+0.53$ (94\%) & $-0.59$ (100\%) \\
Nitro & $+0.34$ (95\%) & $-0.24$ (93\%) \\
Cyano & $+0.19$ (92\%) & $-0.14$ (93\%) \\
\bottomrule
\end{tabular}
\caption{Mean signed attribution of seven common polar functional groups, computed for each regression dataset. Parentheses give the percentage of molecules containing the group in which the attribution sign matches the group's mean.}
\label{tab:mirror}
\end{table}
 
\paragraph{Chemical validity of attributions.} For the functional group view, we average each group's signed attribution over the molecules containing it, separately for ESOL and Lipo, whose targets are aqueous solubility ($\log S$) and lipophilicity ($\log D$). Every group raises predicted solubility and lowers predicted lipophilicity (Table~\ref{tab:mirror}), mirroring the textbook inverse relation captured by the general solubility equation~\cite{jain2001}. This direction holds in 88 to 100 percent of the molecules that contain each group.

Beyond the groups in Table~\ref{tab:mirror}, the carboxylic acid is a clear example. It carries the largest negative lipophilicity attribution in our analysis, a mean of $-1.75$ across 465 molecules in the Lipo dataset. Its near-complete ionization at physiological pH sharply lowers $\log D$~\cite{lassalas2016}. The same ionization suppresses blood-brain barrier permeability~\cite{wager2010}, which the rationale captures in 93 percent of the molecules in the BBBP dataset that contain it. These attributions align with established chemistry, so the evidence the rationale carries is not only readable but chemically grounded.

\begin{figure}[tb]
    \centering
    \includegraphics[width=0.9\columnwidth]{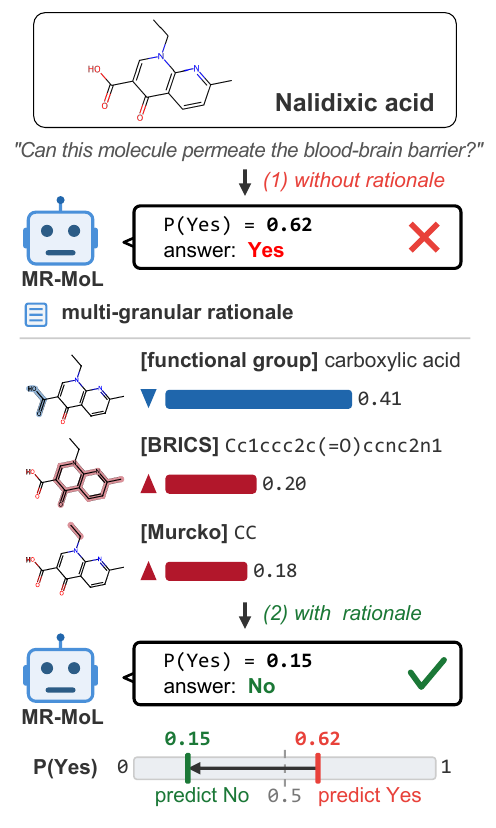}
    \caption{A multi-granular rationale corrects a blood-brain barrier permeability prediction for nalidixic acid. The number beside each substructure in the rationale is its attribution magnitude $|a_{t,j}|$, which ranks the items. The bottom axis shows the resulting shift of $P(\mathrm{Yes})$, the predicted probability of the ``Yes'' label, across the decision boundary.}
    \label{fig:case}
\end{figure}

\begin{table}[t]
\centering
\small
\setlength{\tabcolsep}{5pt}
\begin{tabular}{llcccc}
\toprule
Dataset & Molecule & Truth & No & Full & Rm. \\
\midrule
BACE & HEA inhibitor & Yes & $0.27$ & $0.90$ & $0.25$ \\
ESOL & Perylene & $-8.79$ & $-5.97$ & $-7.10$ & $-3.90$ \\
\bottomrule
\end{tabular}
\caption{Rationale item interventions. Cells show $P(\mathrm{Yes})$ for classification and the predicted value for regression. ``No'' is the prediction without the rationale, ``Full'' uses all rationale items, and ``Rm.'' removes the top-attributed item. ``Truth'' is the ground-truth label or value.}
\label{tab:fixunfix}
\end{table}

\paragraph{Correction of individual predictions.} Moving from populations to individual molecules, we take a molecule that \model{} predicts incorrectly without the rationale, and add the rationale to its input. Figure~\ref{fig:case} traces how \model{} resolves one case in blood-brain barrier permeability (BBBP). Given nalidixic acid without the rationale, \model{} wrongly predicts that it crosses the barrier. In the rationale, the items from the BRICS and Murcko views are tagged as raising permeability, while the functional group carboxylic acid is tagged as lowering it with the largest magnitude. With this evidence, \model{} flips to the correct No.

The same pattern holds on other tasks. Table~\ref{tab:fixunfix} shows one molecule each for BACE and ESOL, where the prediction without the rationale is wrong and the full rationale corrects it. In each case the most influential substructure is one a chemist would point to, and the direction the rationale assigns it matches known chemistry. For Perylene, the fused-ring scaffold is tagged toward lower solubility, consistent with its large hydrophobic surface and stable crystal packing~\cite{jain2001}. In the HEA inhibitor, the hydroxyethylamine scaffold is tagged toward higher activity, consistent with its protonated amine forming a salt bridge with a catalytic aspartate~\cite{maillard2007design}. As a further check, deleting this item returns the model to its error, showing that the correction hinges on that specific item.

\section{Limitations}

We note several limitations that also point to future directions. First, the rationale inherits the quality of the source predictor and may carry misleading evidence. Stronger attribution methods or an ensemble of source predictors could reduce this risk, which we leave to future work. Second, the Murcko, BRICS, and functional group views are not exhaustive, leaving patterns such as macrocycles and stereochemistry-driven properties uncovered. Third, the rationale channel covers only tasks where the GNN source predictor yields attributions, namely classification and regression, leaving tasks such as molecule captioning and reaction prediction beyond its current reach. Extending it to such tasks is another direction for future work.

\section{Conclusion}

We presented \model{}, a molecular LLM that uses explicit, ranked, GNN-derived structural evidence as a rationale alongside the molecular graph and SMILES. Together they let the model predict molecular properties while grounding each prediction in the substructures that drive it. Our diagnostics confirm that the model acts on the rationale's content, not its presence alone, and that the cues it carries reflect known chemistry. Beyond property prediction, this turns model explanations from a diagnostic readout into an input that sharpens the model itself.

{
\small
\bibliography{reference}

\begin{thebibliography}{54}
\providecommand{\natexlab}[1]{#1}

\bibitem[{{Anthropic}(2026)}]{anthropic2026opus46}
{Anthropic}. 2026.
\newblock Claude Opus 4.6 System Card.
\newblock \url{https://www.anthropic.com/claude-opus-4-6-system-card}.
\newblock Accessed: 2026-06-09.

\bibitem[{Bemis and Murcko(1996)}]{bemis1996properties}
Bemis, G.~W.; and Murcko, M.~A. 1996.
\newblock The properties of known drugs. 1. Molecular frameworks.
\newblock \emph{Journal of medicinal chemistry}, 39(15): 2887--2893.

\bibitem[{Cao et~al.(2025)Cao, Liu, Lu, Yao, and Li}]{cao2025instructmol}
Cao, H.; Liu, Z.; Lu, X.; Yao, Y.; and Li, Y. 2025.
\newblock Instructmol: Multi-modal integration for building a versatile and reliable molecular assistant in drug discovery.
\newblock In \emph{Proceedings of the 31st International Conference on Computational Linguistics}, 354--379.

\bibitem[{Chen et~al.(2024)Chen, Yao, Zhang, Cheng, and Bian}]{chen2024hight}
Chen, Y.; Yao, Q.; Zhang, J.; Cheng, J.; and Bian, Y. 2024.
\newblock Hight: Hierarchical graph tokenization for graph-language alignment.
\newblock arXiv:2406.14021.

\bibitem[{Chithrananda, Grand, and Ramsundar(2020)}]{chithrananda2020chemberta}
Chithrananda, S.; Grand, G.; and Ramsundar, B. 2020.
\newblock ChemBERTa: large-scale self-supervised pretraining for molecular property prediction.
\newblock arXiv:2010.09885.

\bibitem[{Degen et~al.(2008)Degen, Wegscheid-Gerlach, Zaliani, and Rarey}]{degen2008art}
Degen, J.; Wegscheid-Gerlach, C.; Zaliani, A.; and Rarey, M. 2008.
\newblock On the art of compiling and using 'drug-like' chemical fragment spaces.
\newblock \emph{ChemMedChem}, 3(10): 1503.

\bibitem[{Edwards, Zhai, and Ji(2021)}]{edwards2021text2mol}
Edwards, C.; Zhai, C.; and Ji, H. 2021.
\newblock Text2mol: Cross-modal molecule retrieval with natural language queries.
\newblock In \emph{Proceedings of the 2021 conference on empirical methods in natural language processing}, 595--607.

\bibitem[{Fang et~al.(2024)Fang, Liang, Zhang, Liu, Huang, Chen, Fan, and Chen}]{fang2024molinstructions}
Fang, Y.; Liang, X.; Zhang, N.; Liu, K.; Huang, R.; Chen, Z.; Fan, X.; and Chen, H. 2024.
\newblock Mol-Instructions: A Large-Scale Biomolecular Instruction Dataset for Large Language Models.
\newblock In \emph{The Twelfth International Conference on Learning Representations}.

\bibitem[{Fang et~al.(2023)Fang, Zhang, Zhang, Chen, Zhuang, Shao, Fan, and Chen}]{fang2023knowledge}
Fang, Y.; Zhang, Q.; Zhang, N.; Chen, Z.; Zhuang, X.; Shao, X.; Fan, X.; and Chen, H. 2023.
\newblock Knowledge graph-enhanced molecular contrastive learning with functional prompt.
\newblock \emph{Nature Machine Intelligence}, 5(5): 542--553.

\bibitem[{Gilmer et~al.(2017)Gilmer, Schoenholz, Riley, Vinyals, and Dahl}]{gilmer2017neural}
Gilmer, J.; Schoenholz, S.~S.; Riley, P.~F.; Vinyals, O.; and Dahl, G.~E. 2017.
\newblock Neural message passing for quantum chemistry.
\newblock In \emph{International conference on machine learning}, 1263--1272. Pmlr.

\bibitem[{Grattafiori et~al.(2024)Grattafiori, Dubey, Jauhri, Pandey, Kadian, Al-Dahle, Letman, Mathur, Schelten, Vaughan et~al.}]{grattafiori2024llama}
Grattafiori, A.; Dubey, A.; Jauhri, A.; Pandey, A.; Kadian, A.; Al-Dahle, A.; Letman, A.; Mathur, A.; Schelten, A.; Vaughan, A.; et~al. 2024.
\newblock The llama 3 herd of models.
\newblock arXiv:2407.21783.

\bibitem[{Hastings et~al.(2012)Hastings, De~Matos, Dekker, Ennis, Harsha, Kale, Muthukrishnan, Owen, Turner, Williams et~al.}]{hastings2012chebi}
Hastings, J.; De~Matos, P.; Dekker, A.; Ennis, M.; Harsha, B.; Kale, N.; Muthukrishnan, V.; Owen, G.; Turner, S.; Williams, M.; et~al. 2012.
\newblock The ChEBI reference database and ontology for biologically relevant chemistry: enhancements for 2013.
\newblock \emph{Nucleic acids research}, 41(D1): D456--D463.

\bibitem[{Hu et~al.(2022)Hu, Shen, Wallis, Allen-Zhu, Li, Wang, Wang, and Chen}]{hu2022lora}
Hu, E.~J.; Shen, Y.; Wallis, P.; Allen-Zhu, Z.; Li, Y.; Wang, S.; Wang, L.; and Chen, W. 2022.
\newblock LoRA: Low-Rank Adaptation of Large Language Models.
\newblock In \emph{International Conference on Learning Representations}.

\bibitem[{Hu* et~al.(2020)Hu*, Liu*, Gomes, Zitnik, Liang, Pande, and Leskovec}]{Hu*2020Strategies}
Hu*, W.; Liu*, B.; Gomes, J.; Zitnik, M.; Liang, P.; Pande, V.; and Leskovec, J. 2020.
\newblock Strategies for Pre-training Graph Neural Networks.
\newblock In \emph{International Conference on Learning Representations}.

\bibitem[{Jain and Yalkowsky(2001)}]{jain2001}
Jain, N.; and Yalkowsky, S.~H. 2001.
\newblock Estimation of the aqueous solubility I: Application to organic nonelectrolytes.
\newblock \emph{Journal of Pharmaceutical Sciences}, 90(2): 234--252.

\bibitem[{Jim{\'e}nez-Luna, Grisoni, and Schneider(2020)}]{jimenez2020drug}
Jim{\'e}nez-Luna, J.; Grisoni, F.; and Schneider, G. 2020.
\newblock Drug discovery with explainable artificial intelligence.
\newblock \emph{Nature Machine Intelligence}, 2(10): 573--584.

\bibitem[{Jin et~al.(2025)Jin, Guo, Zhou, and Guan}]{jin2025effective}
Jin, C.; Guo, S.; Zhou, S.; and Guan, J. 2025.
\newblock Effective and explainable molecular property prediction by chain-of-thought enabled large language models and multi-modal molecular information fusion.
\newblock \emph{Journal of Chemical Information and Modeling}, 65(11): 5438--5455.

\bibitem[{Kim, Lee, and Hwang(2026)}]{kim2026mol}
Kim, D.; Lee, W.; and Hwang, S.~J. 2026.
\newblock Mol-llama: Towards general understanding of molecules in large molecular language model.
\newblock \emph{Advances in Neural Information Processing Systems}, 38: 26921--26960.

\bibitem[{Knox et~al.(2024)Knox, Wilson, Klinger, Franklin, Oler, Wilson, Pon, Cox, Chin, Strawbridge et~al.}]{knox2024drugbank}
Knox, C.; Wilson, M.; Klinger, C.~M.; Franklin, M.; Oler, E.; Wilson, A.; Pon, A.; Cox, J.; Chin, N.~E.; Strawbridge, S.~A.; et~al. 2024.
\newblock DrugBank 6.0: the DrugBank knowledgebase for 2024.
\newblock \emph{Nucleic acids research}, 52(D1): D1265--D1275.

\bibitem[{Lassalas et~al.(2016)Lassalas, Gay, Lasfargeas, James, Tran, Vijayendran, Brunden, Kozlowski, Thomas, Smith, Huryn, and Ballatore}]{lassalas2016}
Lassalas, P.; Gay, B.; Lasfargeas, C.; James, M.~J.; Tran, V.; Vijayendran, K.~G.; Brunden, K.~R.; Kozlowski, M.~C.; Thomas, C.~J.; Smith, I., Amos~B; Huryn, D.~M.; and Ballatore, C. 2016.
\newblock Structure property relationships of carboxylic acid isosteres.
\newblock \emph{Journal of Medicinal Chemistry}, 59(7): 3183--3203.

\bibitem[{Lee et~al.(2026)Lee, De~Brouwer, Hajiramezanali, Biancalani, Park, and Scalia}]{lee2026rag}
Lee, N.; De~Brouwer, E.; Hajiramezanali, E.; Biancalani, T.; Park, C.; and Scalia, G. 2026.
\newblock Rag-enhanced collaborative llm agents for drug discovery.
\newblock In \emph{Proceedings of the AAAI Conference on Artificial Intelligence}, volume~40, 561--569.

\bibitem[{Li et~al.(2023)Li, Li, Savarese, and Hoi}]{li2023blip}
Li, J.; Li, D.; Savarese, S.; and Hoi, S. 2023.
\newblock Blip-2: Bootstrapping language-image pre-training with frozen image encoders and large language models.
\newblock In \emph{International conference on machine learning}, 19730--19742. PMLR.

\bibitem[{Li et~al.(2024)Li, Liu, Luo, Wang, He, Kawaguchi, Chua, and Tian}]{li2024towards}
Li, S.; Liu, Z.; Luo, Y.; Wang, X.; He, X.; Kawaguchi, K.; Chua, T.-S.; and Tian, Q. 2024.
\newblock Towards 3d molecule-text interpretation in language models.
\newblock In \emph{International Conference on Learning Representations}, volume 2024, 17352--17371.

\bibitem[{Liu et~al.(2024{\natexlab{a}})Liu, Ren, Tao, and Ren}]{liu2024git}
Liu, P.; Ren, Y.; Tao, J.; and Ren, Z. 2024{\natexlab{a}}.
\newblock Git-mol: A multi-modal large language model for molecular science with graph, image, and text.
\newblock \emph{Computers in biology and medicine}, 171: 108073.

\bibitem[{Liu et~al.(2022)Liu, Wang, Liu, Lasenby, Guo, and Tang}]{liu2022pretraining}
Liu, S.; Wang, H.; Liu, W.; Lasenby, J.; Guo, H.; and Tang, J. 2022.
\newblock Pre-training Molecular Graph Representation with 3D Geometry.
\newblock In \emph{International Conference on Learning Representations}.

\bibitem[{Liu et~al.(2024{\natexlab{b}})Liu, Ding, Zhou, Fan, and Tan}]{liu2024moleculargpt}
Liu, Y.; Ding, S.; Zhou, S.; Fan, W.; and Tan, Q. 2024{\natexlab{b}}.
\newblock MolecularGPT: Open Large Language Model (LLM) for Few-Shot Molecular Property Prediction.
\newblock arXiv:2406.12950.

\bibitem[{Liu et~al.(2023)Liu, Li, Luo, Fei, Cao, Kawaguchi, Wang, and Chua}]{liu2023molca}
Liu, Z.; Li, S.; Luo, Y.; Fei, H.; Cao, Y.; Kawaguchi, K.; Wang, X.; and Chua, T.-S. 2023.
\newblock Molca: Molecular graph-language modeling with cross-modal projector and uni-modal adapter.
\newblock In \emph{Proceedings of the 2023 Conference on Empirical Methods in Natural Language Processing}, 15623--15638.

\bibitem[{Livne et~al.(2024)Livne, Miftahutdinov, Tutubalina, Kuznetsov, Polykovskiy, Brundyn, Jhunjhunwala, Costa, Aliper, Aspuru-Guzik et~al.}]{livne2024nach0}
Livne, M.; Miftahutdinov, Z.; Tutubalina, E.; Kuznetsov, M.; Polykovskiy, D.; Brundyn, A.; Jhunjhunwala, A.; Costa, A.; Aliper, A.; Aspuru-Guzik, A.; et~al. 2024.
\newblock nach0: multimodal natural and chemical languages foundation model.
\newblock \emph{Chemical Science}, 15(22): 8380--8389.

\bibitem[{Loshchilov and Hutter(2019)}]{loshchilov2018decoupled}
Loshchilov, I.; and Hutter, F. 2019.
\newblock Decoupled Weight Decay Regularization.
\newblock In \emph{International Conference on Learning Representations}.

\bibitem[{Maillard et~al.(2007)Maillard, Hom, Benson, Moon, Mamo, Bienkowski, Tomasselli, Woods, Prince, Paddock et~al.}]{maillard2007design}
Maillard, M.~C.; Hom, R.~K.; Benson, T.~E.; Moon, J.~B.; Mamo, S.; Bienkowski, M.; Tomasselli, A.~G.; Woods, D.~D.; Prince, D.~B.; Paddock, D.~J.; et~al. 2007.
\newblock Design, synthesis, and crystal structure of hydroxyethyl secondary amine-based peptidomimetic inhibitors of human $\beta$-secretase.
\newblock \emph{Journal of medicinal chemistry}, 50(4): 776--781.

\bibitem[{Ouyang et~al.(2022)Ouyang, Wu, Jiang, Almeida, Wainwright, Mishkin, Zhang, Agarwal, Slama, Ray et~al.}]{ouyang2022training}
Ouyang, L.; Wu, J.; Jiang, X.; Almeida, D.; Wainwright, C.; Mishkin, P.; Zhang, C.; Agarwal, S.; Slama, K.; Ray, A.; et~al. 2022.
\newblock Training language models to follow instructions with human feedback.
\newblock \emph{Advances in neural information processing systems}, 35: 27730--27744.

\bibitem[{Park et~al.(2024)Park, Bae, Ko, and Kim}]{park2024llamo}
Park, J.; Bae, M.; Ko, D.; and Kim, H.~J. 2024.
\newblock Llamo: Large language model-based molecular graph assistant.
\newblock \emph{Advances in Neural Information Processing Systems}, 37: 131972--132000.

\bibitem[{Pei et~al.(2025)Pei, Yan, Gao, Zhu, and Wu}]{pei20253d}
Pei, Q.; Yan, R.; Gao, K.; Zhu, J.; and Wu, L. 2025.
\newblock 3D-MolT5: Leveraging Discrete Structural Information for Molecule-Text Modeling.
\newblock In \emph{International Conference on Learning Representations}.

\bibitem[{Ramsundar et~al.(2019)Ramsundar, Eastman, Walters, Pande, Leswing, and Wu}]{Ramsundar-et-al-2019}
Ramsundar, B.; Eastman, P.; Walters, P.; Pande, V.; Leswing, K.; and Wu, Z. 2019.
\newblock \emph{Deep Learning for the Life Sciences}.
\newblock O'Reilly Media.
\newblock \url{https://www.amazon.com/Deep-Learning-Life-Sciences-Microscopy/dp/1492039837}.

\bibitem[{{RDKit}(2006)}]{rdkit}
{RDKit}. 2006.
\newblock RDKit: Open-source cheminformatics.
\newblock \url{https://www.rdkit.org}.
\newblock Accessed: 2026-05-06.

\bibitem[{Ross et~al.(2022)Ross, Belgodere, Chenthamarakshan, Padhi, Mroueh, and Das}]{ross2022large}
Ross, J.; Belgodere, B.; Chenthamarakshan, V.; Padhi, I.; Mroueh, Y.; and Das, P. 2022.
\newblock Large-scale chemical language representations capture molecular structure and properties.
\newblock \emph{Nature Machine Intelligence}, 4(12): 1256--1264.

\bibitem[{Scarselli et~al.(2008)Scarselli, Gori, Tsoi, Hagenbuchner, and Monfardini}]{scarselli2008graph}
Scarselli, F.; Gori, M.; Tsoi, A.~C.; Hagenbuchner, M.; and Monfardini, G. 2008.
\newblock The graph neural network model.
\newblock \emph{IEEE transactions on neural networks}, 20(1): 61--80.

\bibitem[{Sch{\"u}tt et~al.(2018)Sch{\"u}tt, Sauceda, Kindermans, Tkatchenko, and M{\"u}ller}]{schutt2018schnet}
Sch{\"u}tt, K.~T.; Sauceda, H.~E.; Kindermans, P.-J.; Tkatchenko, A.; and M{\"u}ller, K.-R. 2018.
\newblock Schnet--a deep learning architecture for molecules and materials.
\newblock \emph{The Journal of chemical physics}, 148(24).

\bibitem[{Wager et~al.(2010)Wager, Hou, Verhoest, and Villalobos}]{wager2010}
Wager, T.~T.; Hou, X.; Verhoest, P.~R.; and Villalobos, A. 2010.
\newblock Moving beyond rules: the development of a central nervous system multiparameter optimization (CNS MPO) approach to enable alignment of druglike properties.
\newblock \emph{ACS Chemical Neuroscience}, 1(6): 435--449.

\bibitem[{Wang et~al.(2022)Wang, Wang, Cao, and Barati~Farimani}]{wang2022molecular}
Wang, Y.; Wang, J.; Cao, Z.; and Barati~Farimani, A. 2022.
\newblock Molecular contrastive learning of representations via graph neural networks.
\newblock \emph{Nature Machine Intelligence}, 4(3): 279--287.

\bibitem[{Wei et~al.(2022)Wei, Bosma, Zhao, Guu, Yu, Lester, Du, Dai, and Le}]{wei2022finetuned}
Wei, J.; Bosma, M.; Zhao, V.; Guu, K.; Yu, A.~W.; Lester, B.; Du, N.; Dai, A.~M.; and Le, Q.~V. 2022.
\newblock Finetuned Language Models are Zero-Shot Learners.
\newblock In \emph{International Conference on Learning Representations}.

\bibitem[{Wellawatte et~al.(2023)Wellawatte, Gandhi, Seshadri, and White}]{wellawatte2023perspective}
Wellawatte, G.~P.; Gandhi, H.~A.; Seshadri, A.; and White, A.~D. 2023.
\newblock A perspective on explanations of molecular prediction models.
\newblock \emph{Journal of Chemical Theory and Computation}, 19(8): 2149--2160.

\bibitem[{Wu et~al.(2018)Wu, Ramsundar, Feinberg, Gomes, Geniesse, Pappu, Leswing, and Pande}]{wu2018moleculenet}
Wu, Z.; Ramsundar, B.; Feinberg, E.~N.; Gomes, J.; Geniesse, C.; Pappu, A.~S.; Leswing, K.; and Pande, V. 2018.
\newblock MoleculeNet: a benchmark for molecular machine learning.
\newblock \emph{Chemical science}, 9(2): 513--530.

\bibitem[{Wu et~al.(2023)Wu, Wang, Du, Jiang, Kang, Li, Pan, Deng, Cao, Hsieh et~al.}]{wu2023chemistry}
Wu, Z.; Wang, J.; Du, H.; Jiang, D.; Kang, Y.; Li, D.; Pan, P.; Deng, Y.; Cao, D.; Hsieh, C.-Y.; et~al. 2023.
\newblock Chemistry-intuitive explanation of graph neural networks for molecular property prediction with substructure masking.
\newblock \emph{Nature communications}, 14(1): 2585.

\bibitem[{Xia et~al.(2023)Xia, Zhao, Hu, Gao, Tan, Liu, Li, and Li}]{xia2023mole}
Xia, J.; Zhao, C.; Hu, B.; Gao, Z.; Tan, C.; Liu, Y.; Li, S.; and Li, S.~Z. 2023.
\newblock Mole-bert: Rethinking pre-training graph neural networks for molecules.
\newblock In \emph{The Eleventh International Conference on Learning Representations}.

\bibitem[{Xian et~al.(2025)Xian, Gu, Li, and Liang}]{xian2025molrag}
Xian, Z.; Gu, J.; Li, L.; and Liang, S. 2025.
\newblock Molrag: unlocking the power of large language models for molecular property prediction.
\newblock In \emph{Proceedings of the 63rd Annual Meeting of the Association for Computational Linguistics (Volume 1: Long Papers)}, 15513--15531.

\bibitem[{Ying et~al.(2019)Ying, Bourgeois, You, Zitnik, and Leskovec}]{ying2019gnnexplainer}
Ying, Z.; Bourgeois, D.; You, J.; Zitnik, M.; and Leskovec, J. 2019.
\newblock Gnnexplainer: Generating explanations for graph neural networks.
\newblock \emph{Advances in neural information processing systems}, 32.

\bibitem[{Yu et~al.(2024)Yu, Baker, Chen, Ning, and Sun}]{yu2024llasmol}
Yu, B.; Baker, F.~N.; Chen, Z.; Ning, X.; and Sun, H. 2024.
\newblock Lla{SM}ol: Advancing Large Language Models for Chemistry with a Large-Scale, Comprehensive, High-Quality Instruction Tuning Dataset.
\newblock In \emph{First Conference on Language Modeling}.

\bibitem[{Yuan et~al.(2021)Yuan, Yu, Wang, Li, and Ji}]{yuan2021explainability}
Yuan, H.; Yu, H.; Wang, J.; Li, K.; and Ji, S. 2021.
\newblock On explainability of graph neural networks via subgraph explorations.
\newblock In \emph{International conference on machine learning}, 12241--12252. PMLR.

\bibitem[{Zhang et~al.(2021)Zhang, Liu, Wang, Lu, and Lee}]{zhang2021motif}
Zhang, Z.; Liu, Q.; Wang, H.; Lu, C.; and Lee, C.-K. 2021.
\newblock Motif-based graph self-supervised learning for molecular property prediction.
\newblock \emph{Advances in Neural Information Processing Systems}, 34: 15870--15882.

\bibitem[{Zhao et~al.(2023)Zhao, Liu, Ma, Xu, Fu, Deng, Kong, and Liu}]{zhao2023gimlet}
Zhao, H.; Liu, S.; Ma, C.; Xu, H.; Fu, J.; Deng, Z.; Kong, L.; and Liu, Q. 2023.
\newblock GIMLET: A unified graph-text model for instruction-based molecule zero-shot learning.
\newblock \emph{Advances in Neural Information Processing Systems}, 36: 5850--5887.

\bibitem[{Zhao et~al.(2025)Zhao, Zhou, Wang, Fan, Yang, Jiao, Liu, Guo, Lu, Zhou et~al.}]{zhao2025molprophecy}
Zhao, J.; Zhou, Q.; Wang, T.; Fan, Y.; Yang, Q.; Jiao, L.; Liu, C.; Guo, Z.; Lu, Q.; Zhou, F.; et~al. 2025.
\newblock MolProphecy: Bridging medicinal chemists’ knowledge and molecular pre-trained models via a multi-modal framework.
\newblock \emph{Journal of Advanced Research}.

\bibitem[{Zhao et~al.(2024)Zhao, Ma, Chen, Sun, Li, Xia, Xu, Zhu, Zhu, Fan et~al.}]{zhao2024chemdfm}
Zhao, Z.; Ma, D.; Chen, L.; Sun, L.; Li, Z.; Xia, Y.; Xu, H.; Zhu, Z.; Zhu, S.; Fan, S.; et~al. 2024.
\newblock Chemdfm: A large language foundation model for chemistry.
\newblock In \emph{Neurips 2024 Workshop Foundation Models for Science: Progress, Opportunities, and Challenges}.

\bibitem[{Zhou et~al.(2023)Zhou, Gao, Ding, Zheng, Xu, Wei, Zhang, and Ke}]{zhou2023unimol}
Zhou, G.; Gao, Z.; Ding, Q.; Zheng, H.; Xu, H.; Wei, Z.; Zhang, L.; and Ke, G. 2023.
\newblock Uni-Mol: A Universal 3D Molecular Representation Learning Framework.
\newblock In \emph{The Eleventh International Conference on Learning Representations}.

\end{thebibliography}
}

%
\onecolumn
\appendix
\setcounter{secnumdepth}{2} 
\section*{\LARGE Appendix}
\suppressfloats
 
This appendix collects additional detail and analysis that support the main paper. References to the main paper are by section name, and each appendix section below is labeled by a letter for reference from the main text.

\section{Experimental Setup}
\label{app:setup}

\subsection{Dataset Statistics}
\label{app:data}

The eight tasks are MoleculeNet datasets~\cite{wu2018moleculenet}, six binary classification and two regression. BACE asks whether a molecule inhibits human $\beta$-secretase~1, and BBBP whether it penetrates the blood-brain barrier. HIV asks whether a molecule inhibits viral replication. ClinTox has two endpoints, clinical-trial toxicity and FDA approval. Tox21 has 12 toxicity assays, seven nuclear-receptor and five stress-response pathways (NR-AR, NR-AR-LBD, NR-AhR, NR-Aromatase, NR-ER, NR-ER-LBD, NR-PPAR-$\gamma$, SR-ARE, SR-ATAD5, SR-HSE, SR-MMP, SR-p53). SIDER has 27 side-effect classes from the MedDRA system-organ ontology. The regression tasks are ESOL for aqueous solubility and Lipo for the octanol-water distribution coefficient.

Table~\ref{tab:supp-data} reports the per-task split sizes. Every task uses an 8:1:1 scaffold split, following the standard MoleculeNet protocol~\cite{wu2018moleculenet}. For the three multi-label tasks, namely ClinTox, SIDER, and Tox21, we expand each molecule into one query per labeled endpoint, since the model answers a single property at a time. The reported counts are therefore queries, and exceed the number of molecules. The molecule counts, in train/valid/test order, are ClinTox 1{,}149/148/145, SIDER 1{,}141/143/143, and Tox21 6{,}264/783/776. Tox21 departs slightly from the 8:1:1 ratio because unlabeled endpoints are dropped before expansion. The single-label tasks BACE, BBBP, HIV, ESOL, and Lipo have matching query and molecule counts.

\begin{table}[t]
\centering
\small
\begin{tabular}{llrrrrr}
\toprule
Task & Type & End. & Train & Valid & Test & Total \\
\midrule
BACE    & classification & 1  & 1{,}210  & 151   & 152   & 1{,}513  \\
BBBP    & classification & 1  & 1{,}566  & 205   & 194   & 1{,}965  \\
ClinTox & classification & 2  & 2{,}298  & 296   & 290   & 2{,}884  \\
HIV     & classification & 1  & 32{,}901 & 4{,}113 & 4{,}106 & 41{,}120 \\
SIDER   & classification & 27 & 30{,}807 & 3{,}861 & 3{,}861 & 38{,}529 \\
Tox21   & classification & 12 & 63{,}639 & 7{,}151 & 7{,}074 & 77{,}864 \\
ESOL    & regression & 1  & 889      & 109   & 113   & 1{,}111  \\
Lipo    & regression & 1  & 3{,}360  & 420   & 420   & 4{,}200  \\
\bottomrule
\end{tabular}
\caption{Stage~2 dataset statistics. ``Type'' is the prediction task type and ``End.'' is the number of endpoints. Counts are queries.}
\label{tab:supp-data}
\end{table}

\subsection{Reproducibility Details}
\label{app:repro}

We train in two stages, and Table~\ref{tab:supp-config} lists the full hyperparameter setting. The base LLM is Llama-3.1-8B-Instruct. Stage~1 aligns the graph path with the LLM embedding space by training the graph encoder and the projector under a three-group learning rate, while the LLM stays frozen. Stage~2 freezes the graph encoder and the LLM, and adapts the LLM with LoRA~\cite{hu2022lora} jointly with the projector, where the adapters target the attention projections $\{q,k,v,o\}_{\text{proj}}$. The projector holds a Q-Former with eight query tokens followed by a linear projection to the LLM. The effective batch multiplies the per-device batch by the gradient-accumulation steps, giving $2\times32=64$ in Stage~1 and $2\times16=32$ in Stage~2. We select each stage's checkpoint by the lowest held-out validation loss.

\begin{table}[t]
\centering
\small
\setlength{\tabcolsep}{10pt}
\begin{tabular}{lll}
\toprule
 & Stage~1 & Stage~2 \\
\midrule
LLM adaptation      & -- & LoRA ($r{=}16$, $\alpha{=}32$) \\
Trained modules     & graph encoder, projector & LoRA, projector \\
Frozen modules      & LLM & graph encoder, LLM \\
Learning rate       & $1\mathrm{e}{-}5$ / $5\mathrm{e}{-}5$ / $1\mathrm{e}{-}4$ & $1\mathrm{e}{-}4$ / $2\mathrm{e}{-}5$ \\
Optimizer           & AdamW, wd $0.01$ & AdamW, wd $0.01$ \\
Schedule            & cosine, warmup $0.1$ & cosine, warmup $0.05$ \\
Effective batch     & $2\times32=64$ & $2\times16=32$ \\
Max sequence length & $1024$ & $1568$ \\
Epochs              & $20$ & $5$ \\
Gradient clip       & $1.0$ & $1.0$ \\
Precision           & bf16 & bf16  \\
Trainable params    & $99.9$M & $111.7$M \\
\bottomrule
\end{tabular}
\caption{Training configuration for the two stages. The Stage~1 learning rate lists the rates for the graph encoder, the Q-Former, and the linear projection in turn, and the Stage~2 learning rate lists the rates for the LoRA adapters and the projector in turn.}
\label{tab:supp-config}
\end{table}

The randomly initialized projection receives the largest Stage~1 rate, while the pretrained graph encoder receives the smallest. The graph encoder is a five-layer, 300-dimensional GIN. The Q-Former follows the BLIP-2 design~\cite{li2023blip} with a 12-layer SciBERT backbone and six cross-attention layers, all initialized from MolCA~\cite{liu2023molca}. The projection maps the 768-dimensional Q-Former output to the 4096-dimensional LLM hidden size.

\paragraph{Stage~1 checkpoint selection.} We score every Stage~1 checkpoint by next-token prediction loss and perplexity on the standard ChEBI-20 test split of $3{,}300$ molecules, which we hold out as validation data. Table~\ref{tab:supp-stage1} reports the curve. The loss plateaus near epoch~10, so we initialize Stage~2 from that checkpoint.

\begin{table}[t]
\centering
\small
\begin{tabular}{rrr@{\hskip 2em}rrr}
\toprule
Epoch & NTP & PPL & Epoch & NTP & PPL \\
\midrule
1 & 1.174 & 3.236 & 11 & 0.945 & 2.573 \\
2 & 1.120 & 3.065 & 12 & 0.954 & 2.596 \\
3 & 1.062 & 2.891 & 13 & 0.956 & 2.601 \\
4 & 1.041 & 2.831 & 14 & 0.953 & 2.593 \\
5 & 1.011 & 2.749 & 15 & 0.958 & 2.606 \\
6 & 0.993 & 2.700 & 16 & 0.962 & 2.617 \\
7 & 0.993 & 2.699 & 17 & 0.968 & 2.633 \\
8 & 0.982 & 2.671 & 18 & 0.981 & 2.667 \\
9 & 0.950 & 2.587 & 19 & 0.988 & 2.686 \\
\textbf{10} & \textbf{0.943} & \textbf{2.568} & 20 & 0.999 & 2.716 \\
\bottomrule
\end{tabular}
\caption{Stage~1 validation loss across epochs. NTP is the next-token prediction loss and PPL its perplexity. The selected checkpoint is in bold.}
\label{tab:supp-stage1}
\end{table}

\paragraph{Run variability and checkpoint selection.} We repeat the full Stage~2 training three times with random seeds and report the mean and standard deviation, while each baseline is run once. The three runs converge to close validation losses. All diagnostics in the main paper use the run with the lowest validation loss.

\paragraph{Source predictor.} Each task's rationale comes from a Mole-BERT~\cite{xia2023mole} GNN fine-tuned end to end on that task, following the original fine-tuning setup. We use Adam with learning rate $1\mathrm{e}{-}3$, no weight decay, batch size $32$, dropout $0.5$, mean pooling, and $100$ epochs, with no learning-rate schedule. We train each source predictor three times per task, and the rationale uses the checkpoint with the lowest validation loss. The attribution of each item is the masking difference defined in the main paper, and we keep the top five items per molecule.

\paragraph{Hyperparameter selection.} We tune the Stage~2 hyperparameters and select the final configuration by validation loss, as summarized in Table~\ref{tab:supp-hpsearch}.

\begin{table}[t]
\centering
\small
\setlength{\tabcolsep}{5pt}
\begin{tabular}{ll}
\toprule
Hyperparameter & Search range\\
\midrule
LoRA learning rate   & $\{1\mathrm{e}{-}4,\, 5\mathrm{e}{-}5,\, 1\mathrm{e}{-}5\}$\\
LoRA $(r,\alpha)$    & $\{(16,32),(32,64),(8,64),(8,16)\}$\\
Target modules       & attention, or attention $+$ MLP\\
Epochs               & $\{3,5,7\}$\\
Max sequence length  & $\{1024,\, 1568,\, 2048\}$\\
\bottomrule
\end{tabular}
\caption{Stage~2 hyperparameter search. ``Attention'' is the attention projections $\{q,k,v,o\}_{\text{proj}}$ and ``MLP'' the feed-forward projections $\{\text{gate},\text{up},\text{down}\}_{\text{proj}}$.}
\label{tab:supp-hpsearch}
\end{table}

\paragraph{Regression targets.} For ESOL and Lipo we load the MoleculeNet datasets through DeepChem~\cite{Ramsundar-et-al-2019}, which provides the regression targets already z-normalized with training-split statistics. We keep this normalization when fine-tuning the molecular LLMs, namely \model{} and the specialist LLMs, so they generate values on a standardized scale. At evaluation we invert the normalization, and every reported metric is in the original target units. The GNN-based baselines and the source predictor regress the raw values directly.

\paragraph{Environment.} All runs use one NVIDIA RTX~5090 with 32~GB of memory and an Intel Core Ultra 9 285K CPU. The machine runs Ubuntu~24.04.3 LTS with Python~3.12 and CUDA~12.8, and the main libraries are PyTorch~2.11, Transformers~4.57, PEFT~0.19, RDKit~2026.03, and PyTorch Geometric~2.7. Stage~1 takes about $1.5$ hours per epoch, and Stage~2 about six hours per epoch.

\subsection{Prompt Format and Chat Template}
\label{app:prompt}

Both stages format each sample as a single Llama-3.1 chat exchange. Figure~\ref{fig:supp-template} shows the skeleton. The system turn sets the assistant role and, in Stage~2, documents the structure of the rationale. The user turn carries the instruction, the molecular tokens, the SMILES sequence, and the rationale. The assistant turn carries the target, a molecule description in Stage~1 and a property answer in Stage~2. The graph path inserts eight molecule-embedding placeholders between the molecule delimiters, one per Q-Former query, and these are overwritten by the projected molecular embeddings.

\begin{figure}[t]
\centering
\scriptsize
\selectfont
\begin{verbatim}
<|start_header_id|>system<|end_header_id|>
{system_prompt}<|eot_id|>

<|start_header_id|>user<|end_header_id|>
{instruction}
Molecule: <|start_mol_id|> ... <|end_mol_id|>
SMILES: {smiles}
{rationale}<|eot_id|>

<|start_header_id|>assistant<|end_header_id|>
{answer}<|eot_id|>
\end{verbatim}
\caption{Chat template skeleton for both stages. The \texttt{...} between the molecule delimiters stands for eight molecule-embedding placeholders.}
\label{fig:supp-template}
\end{figure}

The Stage~2 system prompt is as follows.

\begin{quote}
``You are a scientific assistant. You will be provided with an instruction, molecular embeddings, the corresponding SMILES string, and an optional GNN-derived structural rationale.

\begin{itemize}\itemsep2pt
  \item The rationale is a ranked list of structural items from three decomposition views, namely Murcko, functional group, and BRICS.
  \item Items are ranked within the same molecule by the magnitude of the GNN prediction change caused by masking each item. Ranks are not globally calibrated and are not compared across molecules.
  \item The effect field describes the direction of the GNN prediction attributed to the substructure, not a ground-truth chemical claim.
  \item For BRICS view, the raw BRICS string and attachments are decomposition metadata and connection labels.
\end{itemize}
When a rationale is provided, treat it as auxiliary evidence, not ground truth.''
\end{quote}

\noindent The Stage~1 system prompt keeps only the role sentence, without the rationale description.

\section{Rationale Construction and Analysis}
\label{app:rationale-part}

\subsection{Source Predictor Choice}
\label{app:sourcepred}
We initialize the source predictor from Mole-BERT~\cite{xia2023mole}. Three reasons motivate this choice. First, Mole-BERT is pretrained to reconstruct masked atoms, so partially masked graphs stay in distribution. This is exactly the regime in which masking attribution operates~\cite{wu2023chemistry}, since each attribution compares the prediction on the full graph with that on a masked one. Second, its GIN backbone matches the graph encoder of our graph path~\cite{liu2023molca}, which keeps the two graph modules architecturally aligned. Third, it is a public and reproducible baseline with a fully specified recipe.

\subsection{Rationale Analysis}
\label{app:rationale}

This section examines what the rationales contain in practice, covering how items distribute across the three views, how the views overlap, the attribution statistics, and a serialized example. For each candidate substructure we compute the masking attribution defined in the main paper, rank items within a molecule by absolute attribution, and keep the top five across the three views.

\paragraph{Granularity composition and overlap.} Table~\ref{tab:supp-gran} reports how the kept items distribute across the three views and how often two items name the same substructure. BRICS contributes the most items overall at $51\%$, followed by Murcko at $43\%$ and functional groups at $6\%$. ESOL is the exception, where the Murcko view dominates. Exact duplication across items is uncommon. Two of the five kept items name an identical substructure in $16.6\%$ of molecules overall, ranging from $0.9\%$ on BACE to $38.6\%$ on ESOL, while three coincide in only $0.5\%$. The views therefore tend to select distinct substructures, so a rationale spans several granularities rather than repeating one fragment. A rationale holds $4.5$ items on average, and $76\%$ of rationales reach the full five.

\begin{table}[t]
\centering
\small
\setlength{\tabcolsep}{5pt}
\begin{tabular}{lccccc}
\toprule
Task & Murcko & BRICS & FG & Dup.\ 2 & Dup.\ 3 \\
\midrule
BACE    & 32.6 & 65.2 & 2.3 & 0.9 & 0.0 \\
BBBP    & 42.9 & 53.2 & 3.8 & 12.8 & 0.4 \\
ClinTox & 40.6 & 53.5 & 5.9 & 15.6 & 0.5 \\
HIV     & 43.7 & 51.2 & 5.0 & 11.3 & 0.3 \\
SIDER   & 41.2 & 52.9 & 5.9 & 17.8 & 0.8 \\
Tox21   & 43.1 & 49.4 & 7.4 & 25.3 & 0.7 \\
ESOL    & 55.9 & 38.7 & 5.4 & 38.6 & 0.7 \\
Lipo    & 36.4 & 59.2 & 4.4 & 10.3 & 0.3 \\
\midrule
Overall & 42.5 & 51.2 & 6.2 & 16.6 & 0.5 \\
\bottomrule
\end{tabular}
\caption{View composition and cross-view duplication of the kept rationale items. The first three columns give the share of items from each view. FG is the functional group view. ``Dup.\ 2'' and ``Dup.\ 3'' are the percentages of molecules in which exactly two and exactly three of the five kept items name an identical substructure.}
\label{tab:supp-gran}
\end{table}

\begin{figure}[t]
\centering
\includegraphics[width=\textwidth]{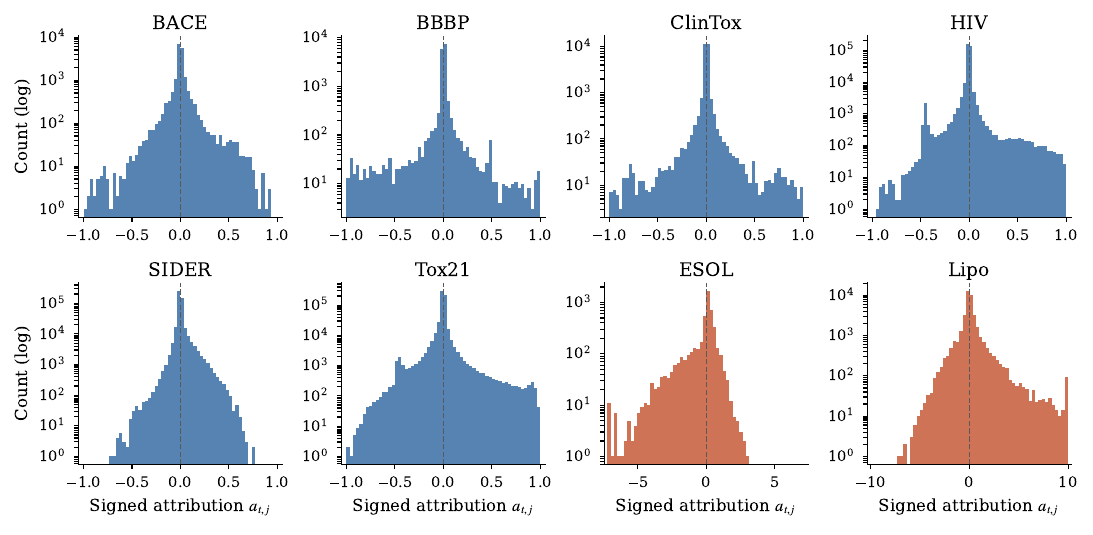}
\caption{Distribution of substructure attributions $a_{t,j}$ per task, on a logarithmic count axis. Classification (blue) is bounded to the probability range $[-1,1]$, regression (orange) is in target units.}
\label{fig:supp-attr}
\end{figure}

\paragraph{Attribution distribution.} Figure~\ref{fig:supp-attr} shows the distribution of signed attributions per task, and Table~\ref{tab:supp-attrstat} gives the summary statistics. Classification attributions lie in $[-1,1]$, since the source predictor outputs a sigmoid probability, so each attribution is a difference of two probabilities. Regression attributions instead carry the target unit and are unbounded, which gives them far larger standard deviations. Every distribution peaks sharply at zero with heavy tails on both sides, so a typical substructure barely moves the source prediction while a few move it sharply. Across tasks the positive share stays between $40$ and $64$ percent, so the predictor assigns both prediction-raising and prediction-lowering roles rather than defaulting to one direction.

\begin{table}[t]
\centering
\small
\setlength{\tabcolsep}{4pt}
\begin{tabular}{lrrrrc}
\toprule
Task & $n$ & Mean & Std & $+\%$ & Range \\
\midrule
BACE    & 18{,}635  & $\phantom{-}0.005$ & 0.120 & 46.8 & $[-0.99,\,0.93]$ \\
BBBP    & 16{,}120  & $-0.003$ & 0.146 & 55.6 & $[-0.99,\,1.00]$ \\
ClinTox & 27{,}028  & $\phantom{-}0.000$ & 0.103 & 50.0 & $[-1.00,\,1.00]$ \\
HIV     & 329{,}838 & $-0.004$ & 0.077 & 45.2 & $[-0.94,\,0.99]$ \\
SIDER   & 465{,}570 & $\phantom{-}0.005$ & 0.052 & 40.4 & $[-0.71,\,0.76]$ \\
Tox21   & 615{,}636 & $-0.004$ & 0.092 & 42.0 & $[-0.98,\,0.99]$ \\
ESOL    & 4{,}796   & $-0.241$ & 1.159 & 63.6 & $[-9.14,\,2.91]$ \\
Lipo    & 39{,}447  & $\phantom{-}0.143$ & 1.330 & 48.4 & $[-7.27,\,16.32]$ \\
\bottomrule
\end{tabular}
\caption{Summary of the attributions $a_{t,j}$. $n$ is the number of candidate substructures, ``Mean'' and ``Std'' the mean and standard deviation of the attributions, ``$+\%$'' the share of positive attributions, and ``Range'' the minimum and maximum.}
\label{tab:supp-attrstat}
\end{table}

\paragraph{Serialized example.} Figure~\ref{fig:supp-example} shows the first three items of one rationale exactly as stored, for one molecule from the ESOL dataset. The three shown items span all three views, one Murcko, one BRICS, and one functional group entry. The Murcko scaffold and the BRICS fragment (ranks~1 and~2) are tagged toward lower predicted solubility, while the nitro group (rank~3) is tagged toward higher. Because the task is regression, the effect field names the predicted value, here the log aqueous solubility for ESOL. For a classification task the field instead names a probability, as in \textit{toward higher predicted probability of blood-brain barrier permeability} for BBBP.

\begin{figure}[tb]
\centering
\scriptsize
\begin{verbatim}
Ranked structural rationale items for this molecule.

1. type: Murcko
scaffold SMILES: O=C(Cn1ccnc1)NCc1ccccc1
effect: toward lower predicted log aqueous solubility

2. type: BRICS
fragment SMILES: O=[N+]([O-])c1nccn1
raw BRICS: [9*]n1ccnc1[N+](=O)[O-]
attachments: [9]
effect: toward lower predicted log aqueous solubility

3. type: functional group
name: nitro
effect: toward higher predicted log aqueous solubility

\end{verbatim}
\caption{The first three items of a serialized rationale as fed to the LLM, for one molecule from the ESOL dataset with SMILES \texttt{O=C(Cn1ccnc1[N+](=O)[O-])NCc1ccccc1}. The full rationale holds five items.}
\label{fig:supp-example}
\end{figure}

\subsection{Number of Rationale Items}
\label{app:numitems}
To test whether supplying multiple items helps, we compare the full five-item rationale with a Top-1 rationale that keeps only the highest-attribution item. The two settings are trained identically, differing only in the number of rationale items the model receives during Stage~2 instruction tuning. Table~\ref{tab:supp-single} reports both on the headline metric. The five-item rationale outperforms the Top-1 setting on most tasks, with the largest gains on BACE and ClinTox, so the lower-ranked items supply complementary signal. HIV is the exception, where the single item does slightly better.

\begin{table}[t]
\centering
\small
\setlength{\tabcolsep}{6pt}
\begin{tabular}{lcccccccc}
\toprule
 & \multicolumn{6}{c}{Classification (ROC-AUC\,$\uparrow$)} & \multicolumn{2}{c}{Regression (RMSE\,$\downarrow$)} \\
\cmidrule(lr){2-7}\cmidrule(lr){8-9}
Variant & BACE & BBBP & ClinTox & HIV & SIDER & Tox21 & ESOL & Lipo \\
\midrule
Top-1 & $80.1_{\pm0.5}$ & $71.3_{\pm1.2}$ & $69.9_{\pm2.6}$ & $\mathbf{74.3}_{\pm0.7}$ & $62.6_{\pm1.4}$ & $72.8_{\pm0.8}$ & $1.216_{\pm0.052}$ & $0.939_{\pm0.013}$ \\
Full  & $\mathbf{82.6}_{\pm0.5}$ & $\mathbf{71.8}_{\pm2.8}$ & $\mathbf{73.8}_{\pm1.4}$ & $72.0_{\pm1.4}$ & $\mathbf{63.3}_{\pm1.3}$ & $\mathbf{74.2}_{\pm0.6}$ & $\mathbf{1.210}_{\pm0.106}$ & $\mathbf{0.919}_{\pm0.006}$ \\
\bottomrule
\end{tabular}
\caption{Comparison of the Full and Top-1 rationales on the headline metric, mean $\pm$ std over three runs. Top-1 keeps only the highest-attribution item, and Full keeps up to five. The better score per task is in bold.}
\label{tab:supp-single}
\end{table}

\section{Additional Results}
\label{app:results}

\subsection{Extended Metrics}
\label{app:extmetrics}
The main paper reports the ablation on ROC-AUC and RMSE. Here we report the corresponding threshold metrics for \model{} and its three channel-ablated variants, the rationale removed (w/o $R$), the graph removed (w/o $G$), and both removed (w/o $G,R$). Classification reports the F1 score, the Matthews correlation coefficient (MCC), and accuracy at the $0.5$ threshold on $P(\mathrm{Yes})$; regression reports the mean absolute error (MAE) and the mean error (ME).

With $TP$, $TN$, $FP$, and $FN$ denoting the four confusion-matrix counts, the F1 score is the harmonic mean of precision and recall,
\begin{equation}
\mathrm{F1} = \frac{2\,TP}{2\,TP + FP + FN},
\end{equation}
and the MCC is
\begin{equation}
\mathrm{MCC} = \frac{TP \cdot TN - FP \cdot FN}{\sqrt{(TP{+}FP)(TP{+}FN)(TN{+}FP)(TN{+}FN)}}.
\end{equation}
F1 ignores $TN$, so it rewards positive-class detection where accuracy does not. MCC ranges over $[-1,1]$ and uses all four counts, so a model that labels every molecule negative scores $0$ rather than the high accuracy such a model attains, which makes MCC the stricter summary on our imbalanced tasks. The MAE is the mean of $|\hat{y}-y|$ and the ME is the signed mean of $\hat{y}-y$, whose sign reveals systematic over- or under-prediction.

Table~\ref{tab:supp-extcls} and Table~\ref{tab:supp-extreg} report these metrics. These threshold metrics expose effects that ROC-AUC alone does not reveal. On ClinTox the two variants without a rationale collapse toward the majority class, with F1 near $48$ and MCC near zero, which mirrors the large ClinTox gain from the rationale. Accuracy alone is misleading on the heavily imbalanced tasks, where w/o $G,R$ reaches high accuracy on ClinTox and HIV yet near-zero MCC, so it labels almost everything negative. Across tasks \model{} attains the best F1 on all tasks but BBBP and the best MCC on all but BBBP and SIDER, indicating that the added channels improve the balanced decision quality.

\begin{table}[tb]
\centering
\small
\setlength{\tabcolsep}{6pt}
\begin{tabular}{lcccccc}
\toprule
Variant & BACE & BBBP & ClinTox & HIV & SIDER & Tox21 \\
\midrule
\rowcolor{gray!20}\multicolumn{7}{l}{\textit{F1 score}} \\
w/o $G,R$  & $37.3_{\pm10.2}$ & $68.2_{\pm0.3}$ & $48.4_{\pm0.0}$ & $2.9_{\pm3.8}$ & $56.3_{\pm0.4}$ & $22.1_{\pm2.6}$ \\
w/o $G$    & $67.8_{\pm3.8}$ & $\mathbf{75.3}_{\pm0.5}$ & $53.2_{\pm5.2}$ & $12.3_{\pm9.5}$ & $59.4_{\pm0.4}$ & $23.1_{\pm3.1}$ \\
w/o $R$    & $64.0_{\pm9.8}$ & $69.4_{\pm1.5}$ & $48.4_{\pm0.0}$ & $17.6_{\pm7.1}$ & $60.3_{\pm2.7}$ & $25.6_{\pm1.8}$ \\
\model{}   & $\mathbf{72.9}_{\pm3.3}$ & $72.5_{\pm2.3}$ & $\mathbf{60.2}_{\pm7.7}$ & $\mathbf{29.0}_{\pm3.5}$ & $\mathbf{63.0}_{\pm0.8}$ & $\mathbf{31.7}_{\pm1.8}$ \\
\midrule
\rowcolor{gray!20}\multicolumn{7}{l}{\textit{MCC}} \\
w/o $G,R$  & $0.225_{\pm.058}$ & $0.154_{\pm.020}$ & $0.000_{\pm.000}$ & $0.059_{\pm.068}$ & $\mathbf{0.103}_{\pm.009}$ & $0.210_{\pm.020}$ \\
w/o $G$    & $0.391_{\pm.049}$ & $\mathbf{0.424}_{\pm.011}$ & $0.151_{\pm.152}$ & $0.183_{\pm.098}$ & $0.056_{\pm.005}$ & $0.207_{\pm.008}$ \\
w/o $R$    & $0.401_{\pm.049}$ & $0.172_{\pm.102}$ & $0.000_{\pm.000}$ & $0.256_{\pm.064}$ & $0.085_{\pm.011}$ & $0.235_{\pm.027}$ \\
\model{}   & $\mathbf{0.491}_{\pm.027}$ & $0.328_{\pm.072}$ & $\mathbf{0.202}_{\pm.138}$ & $\mathbf{0.295}_{\pm.034}$ & $0.101_{\pm.015}$ & $\mathbf{0.299}_{\pm.014}$ \\
\midrule
\rowcolor{gray!20}\multicolumn{7}{l}{\textit{Accuracy}} \\
w/o $G,R$  & $51.5_{\pm4.2}$ & $55.3_{\pm1.6}$ & $\mathbf{93.4}_{\pm0.0}$ & $96.8_{\pm0.0}$ & $75.7_{\pm0.5}$ & $89.8_{\pm0.2}$ \\
w/o $G$    & $67.1_{\pm3.0}$ & $\mathbf{70.6}_{\pm0.5}$ & $92.1_{\pm1.9}$ & $96.9_{\pm0.1}$ & $75.9_{\pm0.1}$ & $88.5_{\pm1.6}$ \\
w/o $R$    & $65.8_{\pm5.6}$ & $57.9_{\pm4.7}$ & $\mathbf{93.4}_{\pm0.0}$ & $\mathbf{97.0}_{\pm0.1}$ & $75.9_{\pm0.4}$ & $89.8_{\pm0.8}$ \\
\model{}   & $\mathbf{72.1}_{\pm2.3}$ & $65.5_{\pm3.6}$ & $90.9_{\pm2.7}$ & $96.7_{\pm0.1}$ & $\mathbf{76.4}_{\pm0.3}$ & $\mathbf{90.2}_{\pm0.3}$ \\
\bottomrule
\end{tabular}
\caption{Extended classification metrics, mean $\pm$ std over three runs. The shaded rows group the results by metric, namely F1 score, MCC, and accuracy. Within each metric block, the best score per task is in bold.}
\label{tab:supp-extcls}
\end{table}

\begin{table}[tb]
\centering
\small
\setlength{\tabcolsep}{5pt}
\begin{tabular}{lcccc}
\toprule
 & \multicolumn{2}{c}{ESOL} & \multicolumn{2}{c}{Lipo} \\
\cmidrule(lr){2-3}\cmidrule(lr){4-5}
Variant & MAE\,($\downarrow$) & ME & MAE\,($\downarrow$) & ME \\
\midrule
w/o $G,R$ & $1.42_{\pm.04}$ & $\phantom{-}0.83_{\pm.14}$ & $0.87_{\pm.02}$ & $-0.06_{\pm.10}$ \\
w/o $G$   & $1.18_{\pm.08}$ & $\phantom{-}0.39_{\pm.29}$ & $0.86_{\pm.01}$ & $\phantom{-}0.23_{\pm.11}$ \\
w/o $R$   & $1.14_{\pm.00}$ & $\phantom{-}0.23_{\pm.23}$ & $0.82_{\pm.01}$ & $\phantom{-}0.02_{\pm.23}$ \\
\model{}  & $\mathbf{0.97}_{\pm.08}$ & $-0.27_{\pm.14}$ & $\mathbf{0.73}_{\pm.01}$ & $\phantom{-}0.07_{\pm.05}$ \\
\bottomrule
\end{tabular}
\caption{Extended regression metrics, mean $\pm$ std over three runs. The lowest MAE per task is in bold.}
\label{tab:supp-extreg}
\end{table}

\subsection{Output Validity}
\label{app:validity}
Validity is the fraction of generated answers on the test split that are parseable. Classification validity is $1.0$ for every molecular LLM by construction, since the answer is read from the positive and negative token logits. We therefore report only regression validity, where a model can emit prose without a parseable number. Table~\ref{tab:supp-valid} shows the result. \model{} and four baselines parse every output, while three generalists miss a small fraction on ESOL.

\begin{table}[tb]
\centering
\small
\begin{tabular}{lcc}
\toprule
Model & ESOL & Lipo \\
\midrule
\multicolumn{3}{l}{\textit{Specialist}} \\
3D-MolT5 & 1.000 & 1.000 \\
HIGHT    & 1.000 & 1.000 \\
\midrule
\multicolumn{3}{l}{\textit{Generalist}} \\
GIMLET       & 1.000 & 1.000 \\
nach0        & 0.938 & 1.000 \\
ChemDFM      & 0.982 & 1.000 \\
LlaSMol      & 1.000 & 1.000 \\
MolecularGPT & 0.982 & 1.000 \\
\midrule
\multicolumn{3}{l}{\textit{Ours}} \\
MR-MoL & 1.000 & 1.000 \\
\bottomrule
\end{tabular}
\caption{Regression output validity.}
\label{tab:supp-valid}
\end{table}

\subsection{Generalist Baseline Task Coverage}
\label{app:coverage}
We evaluate the generalist baselines from their released checkpoints with our own instruction on all eight tasks. Retraining them on our split would turn each released generalist into a task-specific model and defeat the purpose of comparing against generalists, so we keep them as released. As a result, each model has seen a different subset of our eight tasks during its own training, and Table~\ref{tab:supp-coverage} makes this coverage explicit. For every model and task, it marks whether the task was trained ($\bullet$), evaluated zero-shot ($\circ$), or absent ($-$) in that model's own paper. We report this coverage so that scores can be read in context. A trained task may benefit from having appeared in the model's own training data, whereas a zero-shot or absent task is evaluated without such exposure. We therefore avoid drawing conclusions from any single cell, and rely on the overall comparison rather than task-by-task claims against these baselines.

\begin{table}[t]
\centering
\small
\setlength{\tabcolsep}{5pt}
\begin{tabular}{lcccccccc}
\toprule
Model & BACE & BBBP & ClinTox & HIV & SIDER & Tox21 & ESOL & Lipo \\
\midrule
LlaSMol      & $-$      & $\bullet$ & $\bullet$ & $\bullet$ & $\bullet$ & $-$      & $\bullet$ & $\bullet$ \\
ChemDFM      & $\bullet$ & $\bullet$ & $\bullet$ & $\bullet$ & $-$      & $\bullet$ & $-$      & $-$ \\
nach0        & $\bullet$ & $\bullet$ & $-$      & $\bullet$ & $-$      & $-$      & $\bullet$ & $\bullet$ \\
MolecularGPT & $\circ$  & $\circ$  & $-$      & $\circ$  & $-$      & $\circ$  & $\circ$  & $\circ$ \\
GIMLET       & $\circ$  & $\circ$  & $-$      & $\circ$  & $-$      & $\circ$  & $\circ$  & $\circ$ \\
\bottomrule
\end{tabular}
\caption{Task coverage of the generalist baselines. $\bullet$ marks a task in the model's own instruction data, $\circ$ a task held out but reported in the model's paper, and $-$ a task absent from that paper.}
\label{tab:supp-coverage}
\end{table}

\end{document}